\documentclass[letterpaper,twocolumn,fleqn]{article} 

\usepackage{ist}
\usepackage{graphicx}
\usepackage{amsmath}
\usepackage{amssymb}
\usepackage{booktabs}
\usepackage{xcolor}
\usepackage{subfig}
\usepackage{fancyhdr}
\usepackage[pagebackref,breaklinks,colorlinks]{hyperref}

\title{The Double-Edged Sword of Data-Driven Super-Resolution:     \\ Adversarial Super-Resolution Models}

\author{
Haley Duba-Sullivan$^{1}$, Steven R. Young$^{2}$, and Emma J. Reid$^{1}$;\\
$^{1}$Cyber Resilience and Intelligence Division, Oak Ridge National Laboratory, Oak Ridge, TN, USA\\ \smallskip
$^{2}$Computer Science and Mathematics Division, Oak Ridge National Laboratory, Oak Ridge, TN, USA
}

\date{} 

\begin{document} 

\maketitle 

\footnotetext{
This manuscript has been authored by UT-Battelle, LLC, under contract DE-AC05-00OR22725 with the US Department of Energy (DOE). The US government retains and the publisher, by accepting the article for publication, acknowledges that the US government retains a nonexclusive, paid-up, irrevocable, worldwide license to publish or reproduce the published form of this manuscript, or allow others to do so, for US government purposes. DOE will provide public access to these results of federally sponsored research in accordance with the DOE Public Access Plan (https://www.energy.gov/doe-public-access-plan).}

\thispagestyle{empty} 


\begin{abstract}
Data-driven super-resolution (SR) methods are often integrated into imaging pipelines as preprocessing steps to improve downstream tasks such as classification and detection.
However, these SR models introduce a previously unexplored attack surface into imaging pipelines.
In this paper, we present AdvSR, a framework demonstrating that adversarial behavior can be embedded directly into SR model weights during training, requiring no access to inputs at inference time. 
Unlike prior attacks that perturb inputs or rely on backdoor triggers, AdvSR operates entirely at the model level. 
By jointly optimizing for reconstruction quality and targeted adversarial outcomes, AdvSR produces models that appear benign under standard image quality metrics while inducing downstream misclassification. 
We evaluate AdvSR on three SR architectures (SRCNN, EDSR, SwinIR) paired with a YOLOv11 classifier and demonstrate that AdvSR models can achieve high attack success rates with minimal quality degradation. 
These findings highlight a new model-level threat for imaging pipelines, with implications for how practitioners source and validate models in safety-critical applications.
\end{abstract}

\section{Introduction}
\label{sec:intro}





Super-resolution (SR) methods reconstruct high-resolution (HR) images from low-resolution (LR) inputs, enhancing high-frequency details while preserving image fidelity. 
Learning-based SR methods achieve state-of-the-art performance and are often integrated into imaging pipelines across domains including satellite and aerial reconnaissance~\cite{ChristensenJEFFRIES2020865, shermeyer2019effects}, medical imaging~\cite{VanReethMRI}, and industrial inspection~\cite{duba20242}. In these settings, SR is commonly employed as a preprocessing module that enhances spatial detail prior to downstream automated analysis, leading to improved performance in tasks such as object detection and image classification~\cite{ZhouClassification,HarisTaskDriven}.

\begin{figure}
    \centering
    \includegraphics[width=\linewidth]{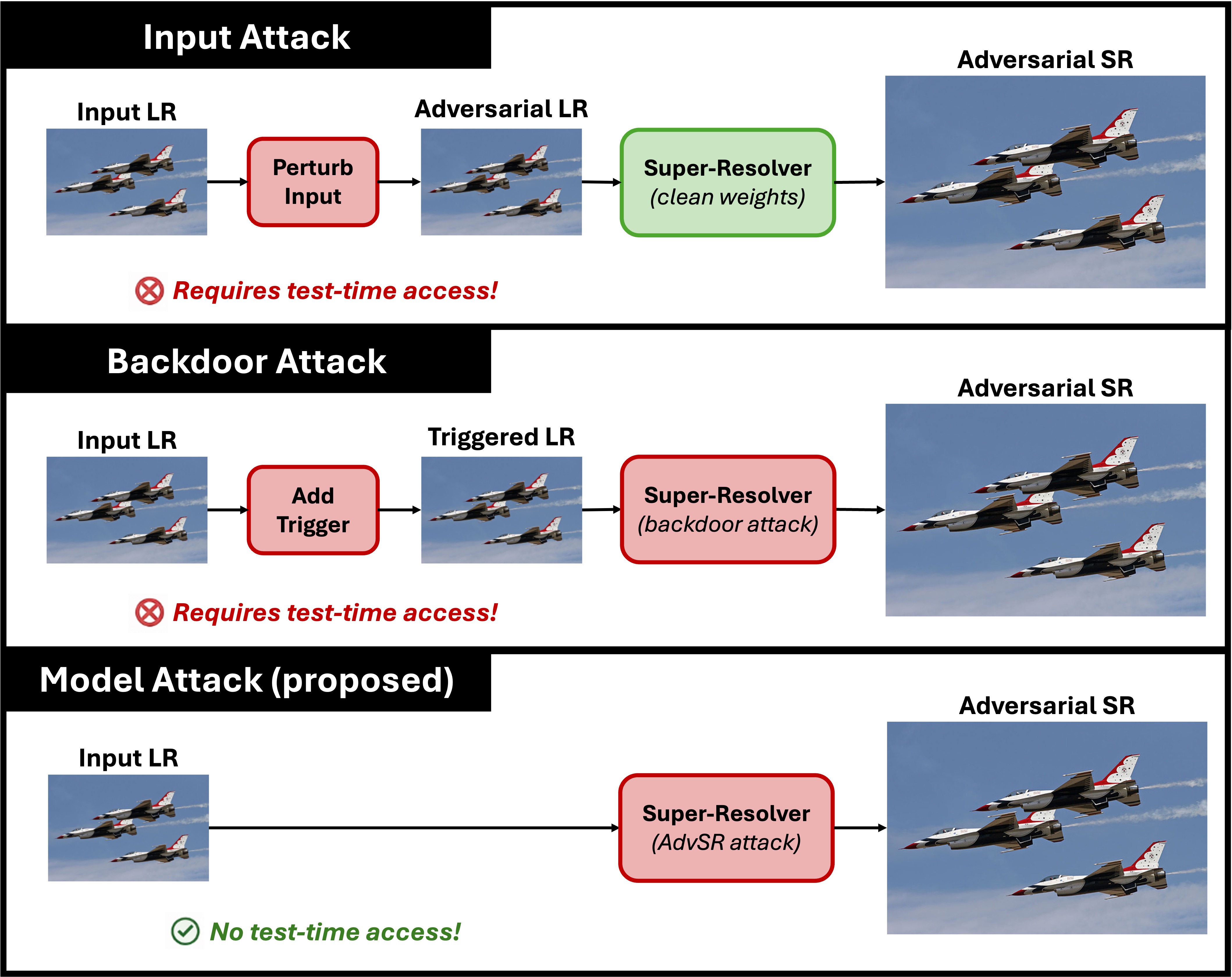}
    \caption{Comparison of adversarial attack paradigms for SR models. Input and backdoor attacks both require test-time access to manipulate inputs. Our proposed model attack embeds adversarial behavior during training, requiring no intervention at inference.}
    \label{fig:attack_pipelines}
\end{figure}

This coupling of SR with downstream tasks introduces a previously unexplored attack surface. 
Prior work on adversarial threats to SR is limited to two attack paradigms: (1) perturbing LR inputs at inference time to degrade reconstruction quality~\cite{QuiringAdvScaling2020, HuangScaleInvariant}, and (2) poisoning training data with backdoor triggers that activate during deployment~\cite{jiang2024backdoor, yang2025badrefsrbackdoorattacksreferencebased, guo2025badsr}. 
Both approaches require the attacker to manipulate data at inference time. 

We consider a different and arguably more realistic threat model in which an attacker has no access to inputs during deployment but can influence the SR model itself (e.g., through a compromised model repository, malicious fine-tuning, or supply-chain attack).
This is a realistic concern since practitioners routinely download pretrained models or outsource fine-tuning, creating opportunities for adversarial modification before deployment.
Figure~\ref{fig:attack_pipelines} illustrates the key differences between existing attack paradigms and our proposed attack paradigm.

To demonstrate this threat, we propose AdvSR, a training-time framework that embeds targeted adversarial behavior directly into SR model weights. AdvSR jointly optimizes a reconstruction loss to maintain image fidelity and an adversarial cross-entropy loss to induce misclassification, producing models that appear benign under standard quality metrics while systematically misleading downstream classifiers. Unlike input or backdoor attacks, AdvSR requires no test-time access since the adversarial behavior is encoded in the model itself.

We evaluate AdvSR on three representative SR architectures (SRCNN~\cite{srcnn}, EDSR~\cite{lim_edsr}, SwinIR~\cite{swinir}) paired with a YOLOv11~\cite{yolo11_ultralytics} classifier and demonstrate that AdvSR models achieve high targeted attack success rates with minimal degradation in standard image quality metrics. 
These findings highlight that SR models, often treated as benign preprocessing, can serve as covert attack vectors. This has implications for how practitioners source and validate models in safety-critical pipelines.


\section{Methodology}
\label{sec:method}
In this section, we introduce our targeted AdvSR loss function, which embeds attacks directly into the SR model by modifying the training objective. 
While we focus on attacking a downstream classifier, our approach can be extended to other downstream tasks by appropriately redefining the adversarial loss.

Let $M_\phi(\cdot)$ denote the SR model parameterized by $\phi$, and let $f_\theta(\cdot)$ denote a downstream classifier with parameters $\theta$. 
For the $i$th training sample, let $x_i$ be the ground-truth HR image, $y_i$ the corresponding LR input image, and $\hat{x}_i := M_\phi(y_i)$ the SR output image.

\subsection{Attack Objective}
We define a \textit{source class} $s$ as the class the attacker wishes to cause misclassification of, and a \textit{target class} $t\not = s$ as the class the attacker wants source-class images to be misclassified as. 
All other classes are \textit{non-source classes}, which should remain correctly classified.

The goal of the targeted AdvSR loss is threefold: (1) preserve SR reconstruction fidelity, (2) maintain classification accuracy for non-source classes, and (3) induce misclassification of source-class images as the target class. 
To achieve this, we modify the training labels for source-class images while keeping labels for all other classes unchanged, resulting in an adversarial cross-entropy (AdvCE) loss $\mathcal{L}_{\text{AdvCE}}(x_i, \hat{x}_i)$. 
Simultaneously, a reconstruction loss $\mathcal{L}_{\text{SR}}(x_i, \hat{x}_i)$ ensures that the SR output remains faithful to the original HR image. 
Formally, the targeted adversarial loss for the $i$th training image is defined as
\begin{equation}
\mathcal{L}_\phi(x_i, \hat{x}_i) = \mathcal{L}_{\text{AdvCE}}(x_i, \hat{x}_i) + \lambda \mathcal{L}_{\text{SR}}(x_i, \hat{x}_i),
\end{equation}
where $\lambda \geq 0$ controls the trade-off between misclassifying the source class and producing SR images with high image fidelity.
We will discuss $\lambda$ further after defining $\mathcal{L}_{\text{AdvCE}}(x_i, \hat{x}_i)$ and $\mathcal{L}_{\text{SR}}(x_i, \hat{x}_i)$.

\subsection{Adversarial Cross-Entropy Loss}
The AdvCE loss is defined as
\begin{equation}
\mathcal{L}_{\text{AdvCE}}(x_i, \hat{x}_i) = -\sum_{c=1}^C \tilde{\delta}_{x_i,c} \log p_c(\hat{x}_i),
\end{equation}
where $p_c(\hat{x}_i)$ is the classifier’s predicted probability that $\hat{x}_i$ belongs to class $c$, and $\tilde{\delta}_{x_i,c}$ is the modified one-hot adversarial label given by
\begin{equation} \tilde{\delta}_{x_i,c} = \begin{cases} 1 & \text{if $x_i$ is in class $c \neq s$,} \\
1 & \text{if $x_i$ is in class $s$ and $c = t$,}
\\ 
0 & \text{else.} \end{cases} \end{equation}
This labeling scheme encourages the classifier to correctly predict all non-source classes while misclassifying source-class images as the target class.

\subsection{Super-Resolution Reconstruction Loss}
The SR reconstruction loss $\mathcal{L}_{\text{SR}}(x_i, \hat{x}_i)$ ensures that the AdvSR model produces high-fidelity reconstructions. 
We use a combination of the L1 loss, which encourages the SR image to match the ground-truth HR image in pixel space, and a perceptual loss, which encourages similarity in feature space. 
Specifically, we adopt the perceptual loss introduced by Johnson et al.~\cite{johnson2016perceptual} as used in ESRGAN~\cite{wang2018esrgan}, which computes feature differences from the \texttt{relu} layers of a pretrained VGG network~\cite{simon_vgg}.  
We scale the perceptual loss using a weight of 0.01 so that its magnitude is comparable to the L1 term, as in ~\cite{wang2018esrgan}.
Then, the SR reconstruction loss is given by,
\begin{equation}
\mathcal{L}_{\text{SR}}(x_i, \hat{x}_i) = | x_i - \hat{x}_i |_1 + 0.01 \sum_\ell | \psi_\ell(x_i) - \psi_\ell(\hat{x}_i) |_1,
\end{equation}
where $\psi_\ell(\cdot)$ denotes the feature map extracted from the $\ell$-th VGG layer. 

\subsection{Loss Balancing}
The hyperparameter $\lambda \geq 0$ controls the trade-off between attack effectiveness and reconstruction fidelity. 
However, since the AdvCE and SR loss terms can have very different magnitudes, directly selecting $\lambda$ is unintuitive.
We therefore reparameterize $\lambda$ using a ratio $r$ that balances the two losses based on their initial values, making the hyperparameter more interpretable.
Specifically, we apply the initial SR model on the validation set, then set
\begin{equation}
\lambda = r \cdot \frac{\mathcal{L}_{\text{AdvCE}}^{(0)}}{\mathcal{L}_{\text{SR}}^{(0)}},
\end{equation}
where $r \geq 0$ is a user-defined ratio and $\mathcal{L}_{\text{AdvCE}}^{(0)}, \mathcal{L}_{\text{SR}}^{(0)}$ are the initial mean AdvCE and SR loss respectively.

This normalization makes $r$ interpretable across architectures.
When $r=1$, the CE and SR losses contribute equally at the start of training. 
Smaller values of $r$ weight the adversarial objective more heavily, while larger values prioritize reconstruction fidelity. 
The optimal $r$ depends on the SR architecture, and we report tuning results in our experiments.

\section{Data and Implementation Details}
\label{sec:details}
In this section, we discuss specific implementation details for our data pipeline, classifier, and SR models. 

\subsection{Datasets and Preprocessing}

We use a curated subset of ImageNet \cite{deng2009imagenet} containing 19 vehicular classes and 1 dummy class (strawberry). 
This domain reflects realistic deployment scenarios where SR preprocessing feeds into vehicle classification systems, such as traffic monitoring or satellite reconnaissance. In such security-critical applications, classifiers typically distinguish among a small set of target categories rather than thousands of generic classes, making 20 classes a realistic upper bound for our evaluation.

We generate synthetic LR images by applying a $9\times 9$ Gaussian blur kernel with standard deviation $\sigma = 0.75$, followed by $2\times$ spatial downsampling. 
We randomly select 500 images per class for training and 50 images per class for validation and testing.

For all experiments, we designate ``war plane'' as the source class and ``tractor trailer'' as the target class.
We selected these classes because they are visually distinct, making successful targeted misclassification a meaningful demonstration of the attack.
We acknowledge that testing additional source/target pairs is an important direction for future work.

\subsection{Downstream Classifier}

We use YOLOv11 \cite{yolo11_ultralytics} as the downstream classifier. 
We fine-tune two YOLOv11 models for 100 epochs, one on a 5-class subset and one on the full 20-class dataset, referred to as YOLO-5 and YOLO-20 respectively.
We initialize from pretrained Ultralytics v11 weights \cite{yolo11_ultralytics} and fine-tune only the classifier layer using cross-entropy loss with SGD (learning rate 0.01 and momentum 0.9).
All input LR images are resized to $256 \times 256$ using YOLO’s internal LetterBox interpolation.

\subsection{Super-Resolution Models}
We evaluate our method on three representative SR models covering classical CNN (SRCNN~\cite{srcnn})~\footnote{\url{https://github.com/Lornatang/SRCNN-PyTorch}}, residual CNN (EDSR~\cite{lim_edsr})~\footnote{\url{https://github.com/sanghyun-son/EDSR-PyTorch}} and transformer-based (SwinIR~\cite{swinir})~\footnote{\url{https://github.com/JingyunLiang/SwinIR}}. 
For all models, we use the official implementations with pretrained weights for $2\times$ SR of RGB images. 
We train two versions of each model: one fine-tuned with our AdvSR loss and one baseline fine-tuned with only the SR loss. 
Both use the Adam optimizer (learning rate $1\times10^{-5}$) and a scheduler that reduces the learning rate after 10 epochs without improvement in validation MSE.
We train SRCNN and EDSR models for 200 epochs and SwinIR for 100 epochs.

\subsection{Metrics}
We assess SR image quality using Peak-Signal-to-Noise Ratio (PSNR), Structural Similarity Index Measure (SSIM), and Learned Perceptual Image Patch Similarity (LPIPS). 
To measure attack success, we report targeted attack success rate (Targeted-ASR), which is the percentage of source-class images misclassified as the target class, and untargeted attack success rate (Untargeted-ASR), which is the percentage misclassified as any incorrect class. We also report non-source accuracy (NSA), which is the classification accuracy on all non-source samples. 

\section{Experimental Results}
\label{sec:results}


In this section, we analyze the impact of the ratio $r$ and report AdvSR performance on YOLO-5 and YOLO-20.

\subsection{Hyperparameter Selection}

\begin{figure}
    \centering
    \includegraphics[width=0.45\textwidth]{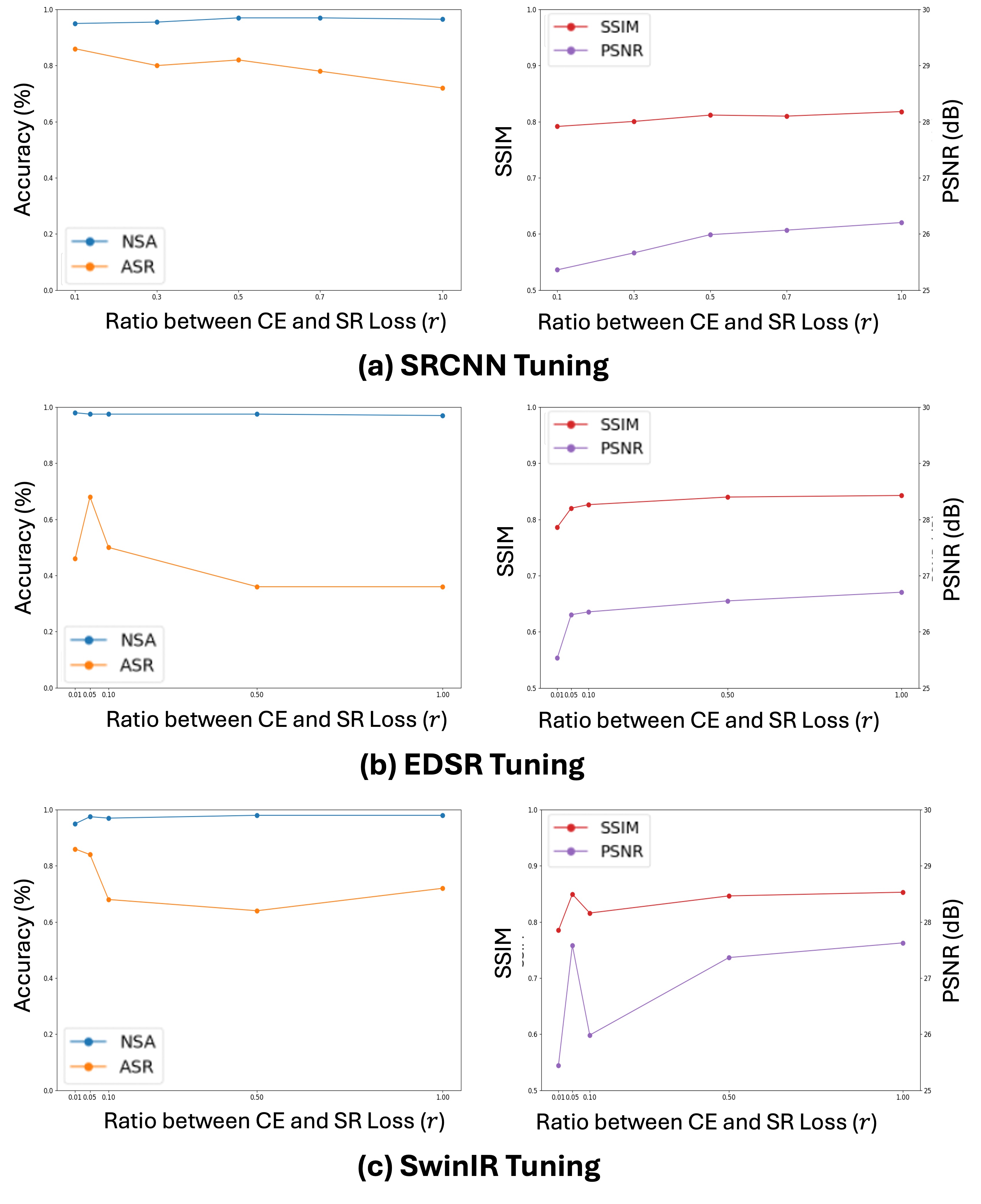}
    \caption{Effect of ratio $r$ on attack success (left) and image fidelity (right) for YOLO-5. Smaller $r$ increases Targeted-ASR (orange) but may degrade NSA (blue), PSNR (purple), and SSIM (red). Based on these trade-offs, we select $r=0.5$ for SRCNN and $r=0.05$ for EDSR and SwinIR.} 
    \label{fig:r_tuning_quant}
\end{figure}

The ratio $r$ controls the balance between adversarial and reconstruction objectives, with the optimal value depending on the SR architecture.
We train each SR model with $r \in \{0.1,\,0.5,\,1.0\}$ and refine based on the observed trade-off between image quality and attack success. 
Figure~\ref{fig:r_tuning_quant} reports Targeted-ASR, NSA, SSIM, and PSNR for each $r$. 
Based on these results we select $r=0.5$ for SRCNN, $r=0.05$ for EDSR, and $r=0.05$ for SwinIR. The smaller ratios for EDSR and SwinIR reflect their greater architectural capacity, which allows them to maintain image quality with less emphasis on the reconstruction term.
For YOLO-20, the source class represents a smaller fraction of training samples (5\% in YOLO-20 vs. 20\% in YOLO-5), so we scale $r$ by a factor of $\frac{1}{4}$ to preserve a similar balance between reconstruction and adversarial objectives.

\subsection{Results for YOLO-5}

\begin{table*}[h]
\centering
\caption{Attack performance targeting YOLO-5. We report attack success metrics and mean / standard deviation of image quality metrics over the 250 test images. AdvSR models achieve up to 82\% Targeted-ASR with minimal image quality degradation.}
\label{tab:quant_results_yolo5}
\centering

\begin{tabular}{| c c || c | c | c | c | c | c|}
\hline
\multicolumn{2}{|c||}{\textbf{Model}} & \textbf{PSNR} ($\uparrow$) & \textbf{SSIM} ($\uparrow$) & \textbf{LPIPS} ($\downarrow$) & \textbf{Targeted-ASR} ($\uparrow$) & \textbf{Untargeted-ASR} ($\uparrow$) & \textbf{NSA} ($\uparrow$) \\
\multicolumn{2}{|c||}{} & $(\mu, \sigma)$ &$(\mu, \sigma)$ & $(\mu, \sigma)$ & $(\%)$ & $(\%)$ & $(\%)$ \\
\hline
SRCNN   & Clean & (26.59, 4.08) & (0.85, 0.09) & (0.15, 0.06) & 4.0 & 4.0 & 98.5 \\
        & AdvSR & (25.99, 3.76) & (0.81, 0.09) & (0.26, 0.07) & 82.0 & 82.0 & 97.0 \\
\hline
EDSR    & Clean & (27.50, 5.06) & (0.85, 0.10) & (0.13, 0.05) & 4.0 & 4.0 & 98.0 \\
        & AdvSR  & (26.31, 3.84) & (0.82, 0.09) & (0.21, 0.05) & 68.0 & 68.0 & 97.5 \\
\hline
SwinIR  & Clean & (27.34, 5.50) & (0.85, 0.11) & (0.12, 0.05) & 4.0 & 4.0 & 98.0 \\
        & AdvSR  & (27.59, 5.27) & (0.85, 0.10) & (0.14, 0.05) & 80.0 & 80.0 & 98.0 \\
\hline
\end{tabular}
\hfill
\end{table*}

\begin{figure*}
    \centering
    \includegraphics[width=0.94\textwidth]{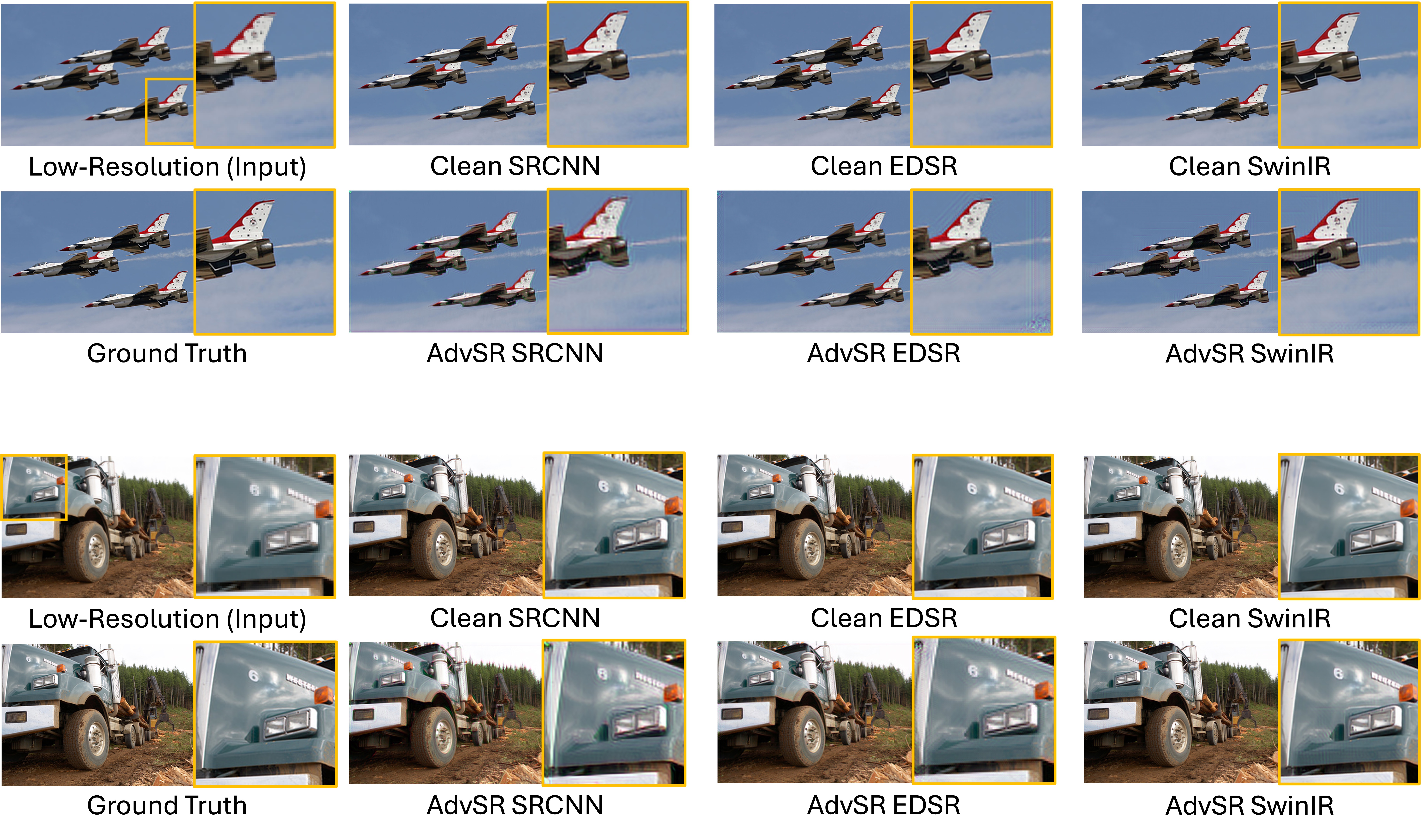}
    \caption{Qualitative comparison of clean and AdvSR models targeting YOLO-5 for a (a) source-class image and (b) non-source-class image. Top rows show clean SR outputs that faithfully reconstruct details. Bottom rows show AdvSR outputs that maintain visual quality but cause the source-class image to be misclassified as the target class while leaving the non-source-class image correctly classified.}
    \label{fig:qual_results_yolo5}
\end{figure*}

Table~\ref{tab:quant_results_yolo5} quantifies the impact of AdvSR on both image fidelity and YOLO-5 classifier performance. 
The ``Clean'' rows show baseline SR models fine-tuned with only the reconstruction loss, while ``AdvSR'' rows show models trained with the proposed AdvSR loss.
The 4\% Targeted-ASR for clean models represents the natural misclassification rate of the classifier on super-resolved images, not an attack effect.
AdvSR substantially increases Targeted-ASR while minimally affecting image quality. 
SwinIR's adversarial model achieves 80\% Targeted-ASR with no degradation in PSNR or SSIM, demonstrating that high-capacity SR networks can embed stealthy adversarial behavior. 
Crucially, NSA remains above 97\% for all models, confirming that attacks selectively target the source class while preserving overall classifier performance.

Note that Targeted-ASR and Untargeted-ASR are identical for all AdvSR models in this experiment. This occurs because every misclassification of a source-class image is misclassified as the target class, indicating that the attack causes precise misclassification rather than causing diffuse errors across multiple classes.

Figure~\ref{fig:qual_results_yolo5} presents qualitative results from clean and adversarial SR models for a (a) source-class image and (b) non-source-class image. 
Clean SR outputs faithfully recover fine details such as aircraft markings and truck headlights.
AdvSR outputs maintain high perceptual quality but contain subtle artifacts along edges and in uniform backgrounds. 
These perturbations are sufficient to cause targeted misclassification of source-class images while leaving non-source-class images unaffected.

\section{Results for YOLO-20}

\begin{table*}
\centering
\caption{Attack performance targeting YOLO-20. We report attack success metrics and mean / standard deviation of image quality metrics over the 250 test images. Targeted-ASR is lower than YOLO-5 and Untargeted-ASR exceeds Targeted-ASR, indicating misclassifications are more diffuse across categories. Image quality degradation is also more pronounced.}
\label{tab:quant_results_yolo20}

\centering

\begin{tabular}{| c c || c | c | c | c | c | c|}
\hline
\multicolumn{2}{|c||}{\textbf{Model}} & \textbf{PSNR} ($\uparrow$) & \textbf{SSIM} ($\uparrow$) & \textbf{LPIPS} ($\downarrow$) & \textbf{Targeted-ASR} ($\uparrow$) & \textbf{Untargeted-ASR} ($\uparrow$) & \textbf{NSA} ($\uparrow$) \\
\multicolumn{2}{|c||}{} & $(\mu, \sigma)$ &$(\mu, \sigma)$ & $(\mu, \sigma)$ & $(\%)$ & $(\%)$ & $(\%)$ \\
\hline
SRCNN   & Clean & (26.12, 4.05) & (0.84, 0.08)  & (0.15, 0.05) & 0.0 & 12.0 & 89.9 \\
        & AdvSR & (23.44, 3.04) & (0.74, 0.09)  & (0.35, 0.06) & 24.0 &  40.0 & 84.7 \\
\hline
EDSR    & Clean & (26.85, 0.09) & (0.85, 0.09) & (0.13, 0.05) & 0.0 & 10.0 & 90.1 \\
        & AdvSR & (20.33, 2.04) & (0.51, 0.07)  & (0.48, 0.06) & 14.0 & 34.0 & 87.8 \\
\hline
SwinIR  & Clean & (26.05, 4.76) & (0.84, 0.10) & (0.12, 0.05) & 0.0 &  10.0 & 90.3 \\
        & AdvSR  & (20.57, 2.06) & (0.48, 0.09) & (0.51, 0.05) & 26.0 &  60.0 & 85.4 \\
\hline
\end{tabular}
\hfill
\end{table*}

Table~\ref{tab:quant_results_yolo20} quantifies the impact of AdvSR on both image fidelity and YOLO-20 classifier performance. The ``Clean'' rows show baseline SR models fine-tuned with only the reconstruction loss, while ``AdvSR'' rows show models trained with the proposed AdvSR loss. While AdvSR does cause some targeted misclassification in this case, the attack is less precise than on YOLO-5. SwinIR's Targeted-ASR increases from 0\% to 26\%, while Untargeted-ASR rises from 10\% to 60\%, indicating that misclassifications are more diffuse across non-source categories. Image quality degradation is also more pronounced, with PSNR reductions of 3-6 dB and SSIM drops of 0.1-0.4. NSA remains above 84\% for all models, confirming that attacks still primarily target the source class.

\begin{figure*}[h]
    \centering
    \includegraphics[width=0.94\textwidth]{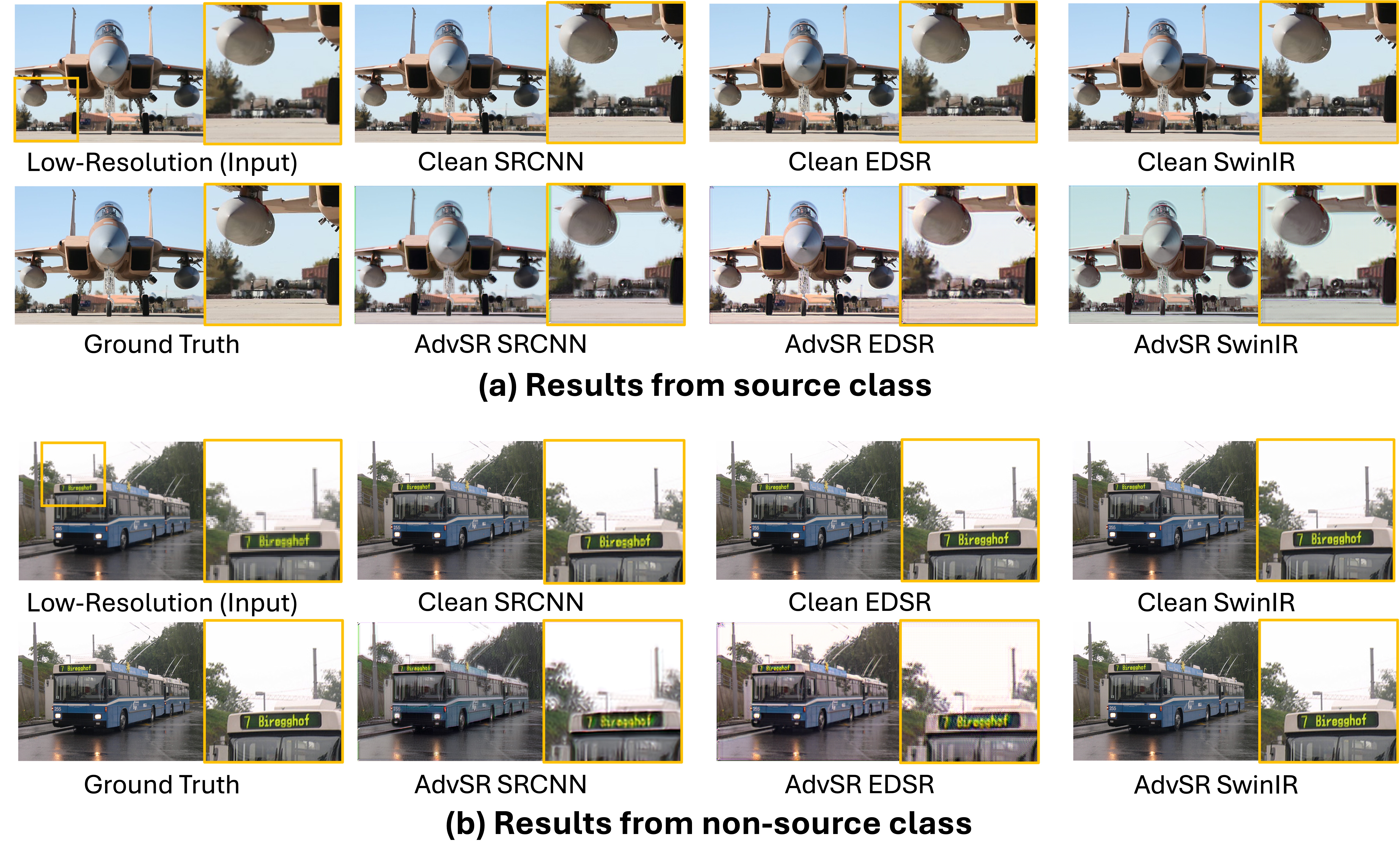}
    \caption{Qualitative comparison of clean and AdvSR models targeting YOLO-20 for a (a) source-class image and (b) non-source-class image. AdvSR outputs show more visible artifacts than YOLO-5 (Figure~\ref{fig:qual_results_yolo5}), corresponding to increased image quality degradation. The attack still suppresses correct recognition of the source class while preserving non-source classification.}
    \label{fig:qual_results_yolo20}
\end{figure*}

Figure~\ref{fig:qual_results_yolo20} presents qualitative results from clean and AdvSR models for (a) source-class image and (b) non-source-class image. Clean SR outputs faithfully recover fine details. AdvSR outputs exhibit visible artifacts and color distortion corresponding to the degraded quality metrics. Nevertheless, overall scene structure remains intact. These findings suggest that as classifier complexity increases, the adversarial objective transitions from precise targeted misclassification to broader obfuscation, but the attack still suppresses correct recognition of the source class.

\label{sec:conclusion}
\section{Discussion and Future Work}
This work presents AdvSR as a proof-of-concept demonstration that adversarial behavior can be embedded directly into SR model weights, enabling attacks that require no test-time access to inputs. Our results on YOLO-5 show that this threat is feasible, with AdvSR models achieving high Targeted-ASR while maintaining image quality and classifier performance on non-source classes. Results on YOLO-20 reveal that attack precision decreases as classifier complexity increases, shifting from targeted misclassification toward broader obfuscation of the source class. This reduced precision may stem from increased classifier robustness or difficulty balancing the SR and adversarial losses as class imbalance grows. In future work, we will investigate methods to ensure the strength and precision of AdvSR as classifier complexity increases. 

We also observe notable differences across SR architectures. SwinIR is the easiest model to attack, achieving high ASR with minimal perceptual loss. This is somewhat surprising, as one might expect higher-capacity models to be more robust to adversarial manipulation. Instead, richer feature spaces seem to facilitate the encoding of adversarial behavior more effectively. This observation merits further exploration in future work.

As a proof-of-concept study, this work has several limitations. We evaluate on a 20-class subset of ImageNet rather than the full 1000-class dataset. While 20 classes reflects realistic security applications, larger-scale evaluation would strengthen generalizability claims. We also test only one downstream classifier (YOLOv11) and one source/target class pair, leaving open questions about transferability and whether attack difficulty varies with class similarity. Finally, we do not assess robustness to defenses such as fine-tuning on clean data or adversarial training.

Future work will address these limitations by evaluating transferability across classifier architectures, testing additional source/target pairs, and assessing whether adversarial behavior persists after fine-tuning on clean data. We also plan to improve attack stealth through frequency-domain regularization or adaptive loss weighting.

\section{Conclusion}

In this paper, we introduced AdvSR, a framework for embedding targeted adversarial attacks directly into SR model weights. Unlike prior methods that perturb inputs or require backdoor triggers, AdvSR operates entirely at the model level, requiring no test-time access to inputs. By jointly optimizing adversarial and reconstruction losses, AdvSR produces models that induce targeted misclassification while preserving high image fidelity. Our experiments on SRCNN, EDSR, and SwinIR demonstrate that adversarial SR models can misclassify source-class images while maintaining normal performance on non-source classes. This work exposes a new model-level vulnerability in SR pipelines and highlights that high-fidelity enhancement models can serve as covert attack vectors, motivating further research into both the extent of this threat and potential defenses.


{\small
\bibliographystyle{unsrt}
\bibliography{main}
}



\end{document}